\documentclass{article} 
\usepackage{colm2024_conference}

\usepackage{microtype}
\usepackage{hyperref}
\usepackage{url}
\usepackage{booktabs}       
\usepackage{amsfonts}       
\usepackage{nicefrac}       
\usepackage{microtype}      
\usepackage{xcolor}         
\usepackage{multirow}

\usepackage{amsmath}
\usepackage{amssymb}
\usepackage{mathtools}
\usepackage{amsthm}

\usepackage{amsfonts}       
\usepackage{nicefrac}       
\usepackage{microtype}      
\usepackage{comment}
\usepackage{enumitem}

\usepackage{colortbl}
\usepackage{graphicx,color}
\usepackage{clrscode}
\usepackage{subfigure}
\usepackage{adjustbox}
\usepackage{multirow}
\usepackage{longtable}
\usepackage{clrscode}
\usepackage{array}
\usepackage{wrapfig}
\usepackage{enumitem}

\usepackage{multirow}
\usepackage{booktabs} 

\title{Position-Aware Parameter Efficient Fine-Tuning Approach for Reducing Positional Bias in LLMs}

\author{\parbox{0.9\linewidth}{
\centering{ Zheng Zhang$^{\dagger}$, Fan Yang$^{*}$, Ziyan Jiang$^{*}$, Zheng Chen$^{*}$, Zhengyang Zhao$^{*}$, Chengyuan Ma$^{*}$, Liang Zhao$^{\dagger}$, Yang Liu\thanks{\rm Amazon, views here are the authors's and not those of Amazon. $^\dagger$Emory University. 
 {\{zheng.zhang,liang.zhao\}@emory.edu} .
 } }
} 
}

\colmfinalcopy 
\begin{document}

\maketitle

\begin{abstract}
Recent advances in large language models (LLMs) have enhanced their ability to process long input contexts. This development is particularly crucial for tasks that involve retrieving knowledge from an external datastore, which can result in long inputs. However, recent studies show a positional bias in LLMs, demonstrating varying performance depending on the location of useful information within the input sequence. In this study, we conduct extensive experiments to investigate the root causes of positional bias. Our findings indicate that the primary contributor to LLM positional bias stems from the inherent positional preferences of different models. We demonstrate that merely employing prompt-based solutions is inadequate for overcoming the positional preferences. To address this positional bias issue of a pre-trained LLM, we developed a \textbf{P}osition-\textbf{A}ware \textbf{P}arameter \textbf{E}fficient \textbf{F}ine-\textbf{T}uning (\textbf{PAPEFT}) approach which is composed of a data augmentation technique and a parameter efficient adapter, enhancing a uniform attention distribution across the input context. Our experiments demonstrate that the proposed approach effectively reduces positional bias, improving LLMs' effectiveness in handling long context sequences for various tasks that require externally retrieved knowledge.
\end{abstract}

\section{Introduction}

Recent advancements in developing Large Language Models (LLMs) significantly enhance the proficiency of language models in harnessing and utilizing extensive input context. This advancement plays a crucial role in improving the performance of applications in areas like recommendation~\citep{naumov2019deep} and question answering~\citep{roberts2020much, yasunaga2021qa}. 
Especially, LLMs have shown remarkable advancements in retrieval-augmented generation tasks, significantly enhancing text information retrieval~\citep{guu2020retrieval,borgeaud2022improving}, exhibiting strong performance in sifting through vast amounts of data to find relevant information. 


\begin{figure}[t]
    \centering
    \includegraphics[width=1.0\textwidth]{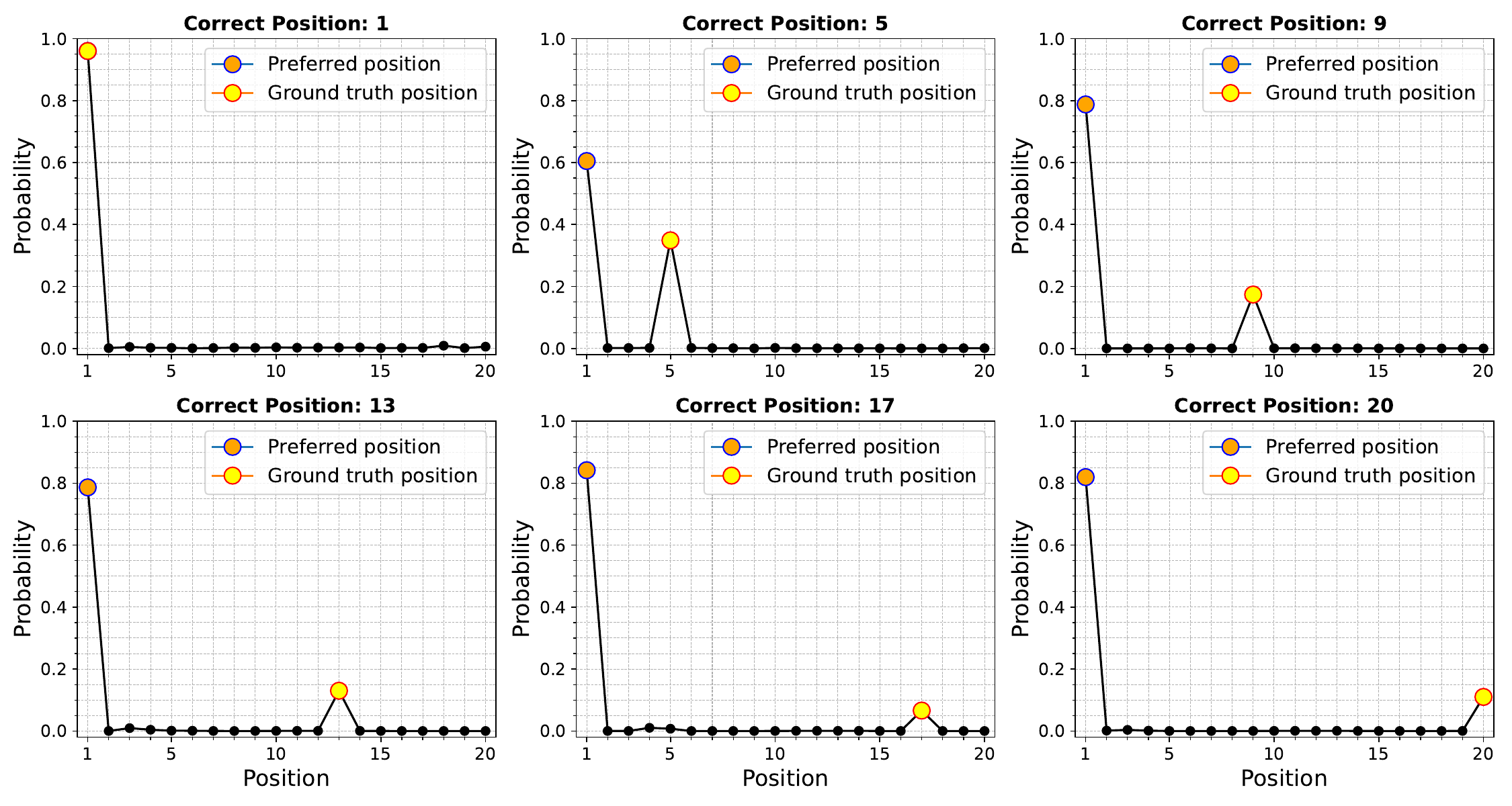} 
    \vspace{-8mm}
    
\caption{Illustration of Positional Preferences in LLMs: The figure demonstrates how the Vicuna-13b-v1.5-16k model's performance on a recommendation task changes with the correct answer's position in the input context window. Given a list of potential candidates, we intentionally position the ground truth candidate at various locations within the list to assess how the predicted position distribution by the LLM shifts. From the figure we can observe the probability peaks near the correct position of relevant information, demonstrating a degree of capacity for identifying pertinent information. There is a notable preference for the first position, indicating significant positional preference.}
\label{fig:prefrence} 
    \vspace{-6mm}
\end{figure}

Although LLMs have made significant progress in processing retrieval-based tasks, their application encounters a key challenge due to a positional bias issue. In many retrieval scenarios, a list of potential candidates is presented. The order of these candidates is often interchangeable and not intended to influence the outcome. However, the inherent input structure of LLMs necessitates flattening this list, thereby imposing an artificial ``ordering'' over the candidates. Recent studies~\citep{liu2023lost, ravaut2023position} have revealed that the performance of LLMs is notably affected by the position of relevant information within the input context, especially in cases of extended input lengths. Specifically, previous study~\citep{liu2023lost} claimed that LLMs often perform better when relevant information is at the beginning or end of a sequence, while their performance decreases when key details are in the middle. The uneven performance across text segments is described as ``lost-in-the-middle'' phenomenon.

While preliminary research~\citep{liu2023lost, ravaut2023position,zheng2023large} has highlighted this positional bias as a significant limitation in LLMs, there is a notable gap in understanding the underlying causes of this issue. In our study, we have conducted comprehensive experiments to assess how the position of genuinely relevant information influences the probability distribution of the retrieved information's location. Our findings indicate that rather than the ``lost-in-the-middle'' phenomenon, it is more accurate to state that each LLM exhibits a unique ``positional preference'' within the context window. For example, as shown in Figure~\ref{fig:prefrence}, the Vicuna-13b-v1.5-16k model exhibits a clear ``positional preference'' for selecting the initial position within the input context as the predicted position, regardless of the actual location of the relevant information.

Moreover, a general while efficient solution to mitigate this positional bias issue remains under-explored. Addressing this challenge is crucial for the advancement and accuracy of LLM applications, especially in contexts where the order of information should not affect the understanding ability of LLMs. Initially, we execute a series of experiments demonstrating that merely employing prompt-based strategies, such as presenting few-shot examples or instructing LLMs to organize candidates hierarchically, can not overcome the issue. To counter this, we introduce a data augmentation strategy that involves permuting the position order within documents to mitigate the positional preference issue inherent in the source data. Additionally, we propose a parameter-efficient fine-tuning technique named \textbf{P}osition-\textbf{A}ware \textbf{P}arameter \textbf{E}fficient \textbf{F}ine-\textbf{T}uning (\textbf{PAPEFT}), designed to make pre-trained LLMs aware of and adjust for positional bias by explicitly considering document positions within the context window. Experimental results across various applications, including recommendation and link prediction, show an over 56\% reduction in performance variance across different positions of relevant information, demonstrating a more consistent and reliable understanding abilities of proposed method within the input context window.

The remainder of this work is organized as follows: We begin by discussing existing relevant studies in the Section~\ref{sec:related_works}. This is followed by a formal problem definition and an introduction to the datasets in the Section~\ref{sec:preliminary}. Subsequently, in the Section~\ref{section:empirical}, we investigate the underlying cause of LLMs' positional bias through a series of empirical studies and shows that simply adopting prompt-based solution can not address the bias. In Section~\ref{sec:method}, we delve into the motivation behind our approach and discuss the specific techniques employed. We conclude with comprehensive experimental results, assessing aspects such as effectiveness and efficiency in Section~\ref{sec:results}.

\section{Related Works}\label{sec:related_works}

\noindent\textbf{Positional Bias in LLMs}. While LLMs have gained prominence, the exploration of positional bias within these models is still in its infancy and has only recently started to attract attention. The body of existing research, though growing, remains limited. A handful of early studies have started to shed light on the implications of positional bias in LLMs. For instance, Liu et al.~\citep{liu2023lost} provided benchmarks indicating that positional bias is a widespread concern, particularly in question answering and key-value pair retrieval tasks. Ravaut et al.~\citep{ravaut2023position} expanded on this by exploring positional bias in text summarization tasks. Zheng et al.~\citep{zheng2023large} made an early attempt to correct this bias by adjusting LLM outputs based on a prior probability reflecting the model's option preferences. Nevertheless, their approach is confined to multiple-choice contexts and lacks generalizability for broader applications. This limitation stems from the challenges in calculating the prior probability and the significant increase in computational demands that such a method entails.

\noindent\textbf{Retrieval-Augmented Generation}. Retrieval-Augmented Generation combines generative capabilities of language models with external knowledge retrieval, enhancing accuracy and relevance in responses. Early foundational work in transformers set the stage for RAG systems \citep{vaswani2017attention}. Subsequent developments like R-Transformer~\citep{lewis2020retrieval} and RAG models~\citep{guu2020realm} integrated retrieval mechanisms with large language models, improving performance in knowledge-intensive tasks. Recent advancements~\citep{liu2023pre, chowdhery2023palm} focus on optimizing retrieval efficiency and accuracy, addressing challenges in coherence, factual correctness, and bias management. 

\noindent\textbf{Parameter Efficient Fine-Tuning for LLMs}.
Parameter-efficient fine-tuning has emerged as a crucial technique for enhancing model performance without the substantial computational and memory costs associated with full model training. This approach, as discussed in recent literature, involves adjusting a small subset of the model's parameters while keeping the majority of the model's weights fixed, thereby enabling the model to adapt to new tasks or data with minimal resource expenditure. Techniques such as adapter layers~\citep{houlsby2019parameter, dettmers2024qlora}, prompt tuning~\citep{li2021prefix}, and sparse updates~\citep{liu2023deja} have been highlighted as effective means for achieving this efficiency. Such methods not only conserve resources but also mitigate the risk of overfitting by limiting the degree of freedom during the training process. Existing research primarily concentrates on enhancing the efficiency of the fine-tuning stage for LLMs, whereas our work is distinctly focused on debiasing a pre-trained LLM using an efficient fine-tuning module.

\section{Preliminaries}\label{sec:preliminary}

\paragraph{Problem Formulation.} This paper focuses on the exploration and analysis of tasks which leverage Retrieval-Augmented Generation (RAG) framework in the context of Large Language Models (LLMs). The central scenario of our study is an input context with a set of $K$ retrieved documents. Among these documents, only one contains the correct or relevant information. The input context to LLM is a composite of all $K$ potential documents alongside additional textual cues. Formally, the input context is structured as $[\mathbf{P}_s, \mathbf{X}_1, \mathbf{X}_2, \dots, \mathbf{X}_K]$, where $\mathbf{P}_s$ represents the additional textual prompts that describe the task for LLMs. The desired outcome from the LLM in response to this input should correctly identify and select the correct relevant document from the set of $K$ candidates. 
Formally, if the correct relevant document is denoted as $\hat{\mathbf{X}}_c$, where $c$ ranges from 1 to $K$, the likelihood that the LLM identifies the $i$-th position as the true relevant location is expressed as $P(\mathbf{X}_i| \hat{\mathbf{X}}_c)$. The fluctuation in the accuracy of predictions for the correct positions, as these positions vary, is measured by the standard deviation divided by the mean of the set of probabilities for all correctly predicted positions. This set is represented as $\{P(\mathbf{X}_c| \hat{\mathbf{X}}_c) | c \in [1, K]\}$, where $P(\mathbf{X}_c| \hat{\mathbf{X}}_c)$ denotes the probability of the correct position $\mathbf{X}_c$ given the predicted position $\hat{\mathbf{X}}_c$, for each position $c$ within the range of 1 to $K$.

\paragraph{Datasets.} Our research investigates positional bias in Language Learning Models (LLMs) across Recommendation (REC), and Link Prediction (LP) domains. We employed specialized datasets—Amazon M2~\citep{jin2023amazon} for REC, and Arxiv~\citep{wang2020microsoft} for LP—to evaluate LLMs' ability to identify key information in varying contextual placements. The critical information's position within each dataset was systematically varied to test LLM adaptability and accuracy with shifting contexts. The details of used datasets are as follows:

\textsc{Recommendation (REC)}: We utilized the Amazon M2 dataset~\citep{jin2023amazon}, which is a rich source of user-product interaction data. Each session within the dataset consists of a sequence of products previously purchased by a user, and their next following purchase. The dataset provides extensive metadata for each product, including descriptions and brand information. We present a set of $K$ possible products per session, among which only one is the actual product that the user purchased and the others are negative samples. We alter the position of ``ground truth'' product within the list to examine the LLMs' proficiency in pinpointing the relevant information depending on its contextual placement.

\textsc{Link-Prediction (LP)}: We leverage the comprehensive citation network benchmark dataset Arxiv~\citep{kwiatkowski2019natural}. This dataset describes a large citation graph where each node is a research paper and the connections between nodes indicate their citation behavior. To assess the ability of LLMs against varied positions of relevant information, we include an evaluation of their ability to accurately identify and present correct cited paper. We manipulate the location of the ground truth cited paper among a list of randomly sampled papers. Specifically, for each given paper, we present a list of papers while only one of them is truly cited by the given one. Then we vary the position of the `ground truth' paper to evaluate how the position of the paper impacts the LLMs' prediction ability.

\paragraph{Choice of LLMs.} 
To assess the robustness of LLMs against positional bias in scenarios involving extensive context sizes, we have compiled a list of widely recognized open-source LLMs specifically tailored for managing long input contexts. This compilation features models frequently employed in academic research, such as Vicuna-13b-v1.5-16k~\citep{vicuna2023} and Longchat-13b-16k~\citep{longchat2023}. These selections enable us to examine a model's efficacy in processing extended dialogues and its capability to handle positional information within conversational settings.

\paragraph{Evaluation Metrics.} 
We adopt ``accuracy'' to evaluate the generated answer quality by LLMs, judging whether the correct relevant document is selected to generate the final answer. Additionally, we employ ``fluctuation'' as a metric to assess the variance in performance across different positions, which is defined as the ratio of the standard deviation to the average value.

\section{Empirical Studies}~\label{section:empirical}\vspace{-10mm}
\begin{figure}[t]
    \centering
    \includegraphics[width=1.0\textwidth]{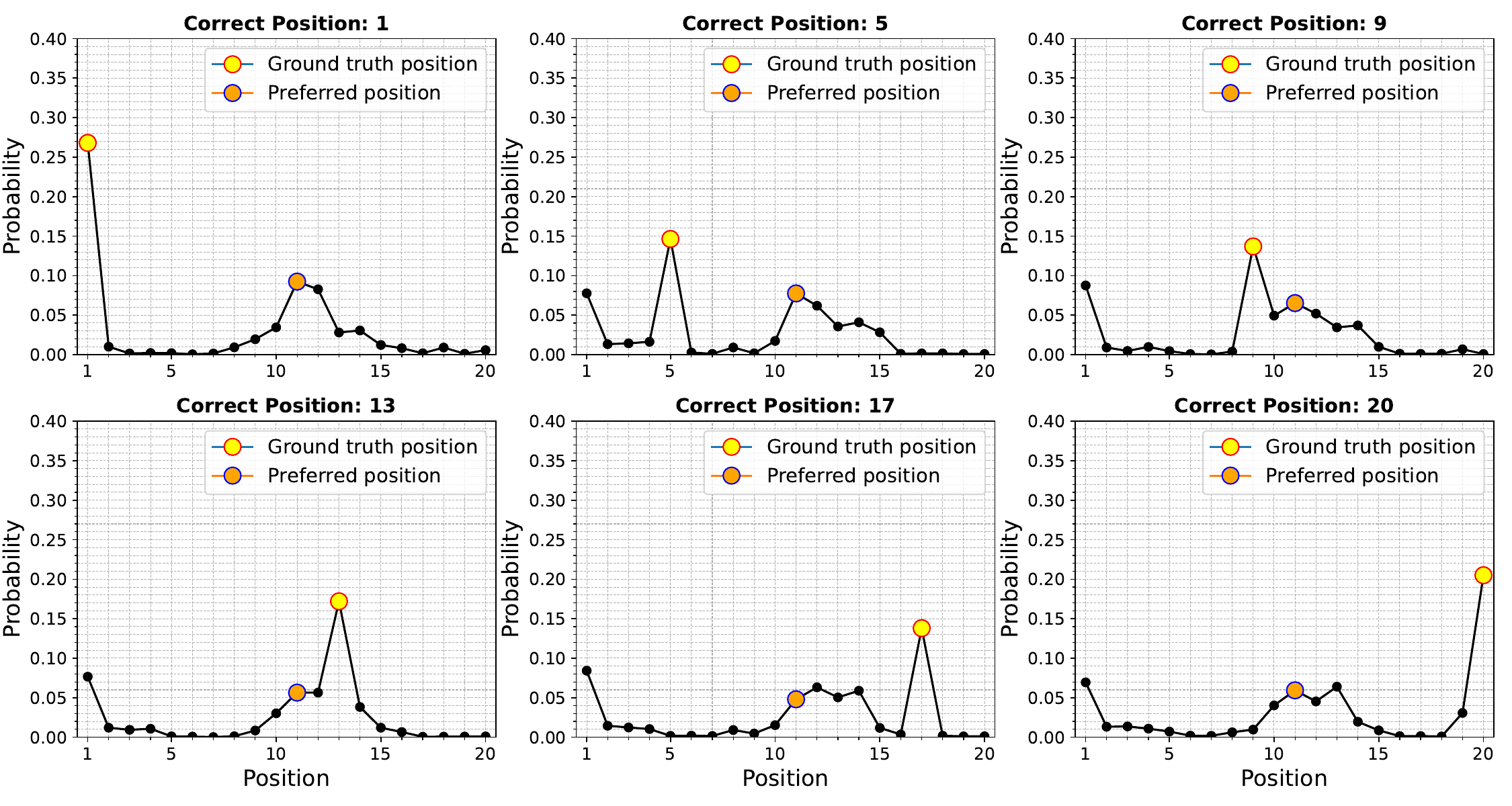} 
    \vspace{-10mm}
    
\caption{The Longchat-13b-16k model's performance on a recommendation task changes with the correct answer's position in the input context window. Comparing with the Vicuna-13b-v1.5-16k model's trend in Figure~\ref{fig:prefrence}, we can observe that these two models have different preferred positions. The Longchat-13b-16k model has preferred location around position eleven, while Vicuna-13b-v1.5-16k prefers the first position.}
\label{fig:prefrence-2} 
\vspace{-3mm}
\end{figure}
\subsection{Investigating the Underlying Causes of Positional Bias in LLMs}
To uncover the reasons underlying positional bias in LLMs, we initiate our investigation by conducting empirical experiments. These experiments are designed to evaluate how the placement of the ground truth answer influences the probability distribution of the positions predicted by LLMs. In Figure~\ref{fig:prefrence} and Figure~\ref{fig:prefrence-2}, we illustrate the predicted probability distributions for all potential positions across various ground truth locations, using the Vicuna-13b-v1.5-16k and Longchat-13b-16k models, respectively. From these figures, it is evident that both models exhibit a ``preferred position'' for the predicted answer, regardless of the actual ground truth positions. Notably, the models demonstrate distinct positional preferences, where Longchat-13b-16k shows a preference for the eleventh position and Vicuna-13b-v1.5-16k tends to favor the first position. Therefore, instead of the ``lost-in-the-middle'' phenomenon suggested by earlier research~\citep{liu2023lost}, we arguably propose that the issue of positional bias is primarily due to the model's ``preferred position''.

\subsection{Prompt-Engineering Based Method Performance}

\begin{table}[t]
\centering
\begin{adjustbox}{width=.75\columnwidth,center}
    \begin{tabular}{l|c|c|c|c|c|c|c}
        \toprule
        Model & \# of Few-shot & 1 & 5 & 9 & 13 & 17 & 20 \\
        \midrule
        Longchat-13b-16k & 0 & 0.426 & 0.125 & 0.105 & 0.142 & 0.127 & 0.261 \\
        Longchat-13b-16k & 1 & 0.620 & 0.223 & 0.167 & 0.161 & 0.150 & 0.338 \\
        Longchat-13b-16k & 3 & 0.594 & 0.169 & 0.130 & 0.142 & 0.123 & 0.269 \\
        Longchat-13b-16k & 5 & 0.669 & 0.175 & 0.114 & 0.108 & 0.100 & 0.228 \\ \hline
        Vicuna-13b-v1.5-16k & 0 & 0.931 & 0.207 & 0.076 & 0.057 & 0.019 & 0.069 \\
        Vicuna-13b-v1.5-16k & 1 & 0.832 & 0.014 & 0.005 & 0.003 & 0.000 & 0.001 \\
        Vicuna-13b-v1.5-16k & 3 & 0.872 & 0.002 & 0.001 & 0.002 & 0.005 & 0.011 \\
        Vicuna-13b-v1.5-16k & 5 & 0.903 & 0.003 & 0.001 & 0.003 & 0.002 & 0.004 \\
        \bottomrule
    \end{tabular}

\end{adjustbox}
    \vspace{-4mm}
    \caption{Few-shot performance on recommendation task with 1, 3, and 5 few-shot examples. Here ``0'' few-shot examples denotes the performance of zero-shot situation. }
    \label{table:fewshot-performance} 
    \vspace{-4mm}
\end{table}

Considering the identified ``preferred position'' bias, a natural question arises: can we devise an effective method to mitigate this bias issue? With recent advancements demonstrating that LLMs possess a significant in-context learning capability~\citep{min2021metaicl}, enabling them to learn and reason based on the text prompts provided, it naturally leads to the question whether specially crafted prompts could be employed to address or alleviate the impact of positional bias. To answer this question, we have crafted a variety of input prompts, aiming to provide insights on addressing the positional bias issue. The description of prompts is as below and the example of prompts can be found in Appendix Table~\ref{zeroshot-prompt}.

(1) \textbf{Zero-shot learning}: LLMs are tasked with generating responses without any prior examples. This setting is essential to observe the natural inclinations of LLMs and their raw handling of positional information, providing a baseline for their performance.

(2) \textbf{Few-shot learning}: We provide the LLMs with a handful of selected examples within the prompt. The goal is to determine if a limited number of illustrative examples can provide sufficient knowledge to the models, thereby guiding them towards more accurate interpretations of information, regardless of its positional context.

(3) \textbf{Hierarchical inference}: A potential cause of the positional bias issue could be attributed to the extensive context size and the large number of possible choices. To tackle this challenge, we suggest employing a prompt that encourages the LLM to make prediction in a bottom-up manner. Initially, the model is instructed to categorize all candidates into a few smaller groups, followed by identifying the most likely answer within each group. Subsequently, from the chosen answers for each group, the model is tasked with making the final prediction. Thus, the overall prediction process is structured in a hierarchical fashion, aiming to mitigate the effects of positional bias.

\subsubsection{Few-shot Learning}
In Table~\ref{table:fewshot-performance}, we present the impact of utilizing a varying number of few-shot examples on the performance in a recommendation task. The findings indicate that while few-shot examples can generally enhance the model's accuracy, they do not mitigate the issue of positional bias along the sequence. The fluctuation in performance across different positions continues to exhibit high variance, even as the quantity of few-shot examples is increased.

\subsubsection{Hierarchical Inference}
The outcomes of employing hierarchical inference are detailed in Table~\ref{table:hierarchical-performance}. It is observed that this approach not only fails to mitigate the positional bias issue but also leads to a notable decrease in performance. A possible explanation for this result could be that LLMs might not effectively process the complex instructions presented within a single input prompt.

In conclusion, the empirical studies showcased here demonstrate that prompt-based solutions alone are insufficient to resolve the positional bias issue. Given the observation of a distinct ``preferred location'' for each pre-trained LLM, it is arguably possible that positional biases are inherently introduced during the pre-training phase or the instruction fine-tuning phase through the training data.

\begin{table}[t]
\centering
\begin{adjustbox}{width=.8\columnwidth,center}
    \begin{tabular}{l|c|c|c|c|c|c|c}
        \toprule
        Model & Hierarchical & 1 & 5 & 9 & 13 & 17 & 20 \\
        \midrule
        Longchat-13b-16k & No & 0.426 & 0.125 & 0.105 & 0.142 & 0.127 & 0.261 \\
        Longchat-13b-16k & Yes & 0.131	& 0.032& 	0.022& 	0.028	& 0.069& 	0.149 \\ \hline
        Vicuna-13b-v1.5-16k & No & 0.931 & 0.207 & 0.076 & 0.057 & 0.019 & 0.069 \\
        Vicuna-13b-v1.5-16k & Yes & 0.179	 & 0.012	 & 0.041	 & 0.280 & 	0.212	 & 0.268 \\ \hline
        GPT-3.5-turbo-16k & No & 0.440	& 0.507	& 0.495& 	0.354	& 0.315 & 	0.288 \\
        GPT-3.5-turbo-16k & Yes & 0.245	& 0.147 & 	0.133& 	0.092	&0.095 &	0.138 \\ 
        \bottomrule
    \end{tabular}

\end{adjustbox}
    \vspace{-4mm}
    \caption{Hierarchical inference performance on recommendation task with varying positions. }
    \label{table:hierarchical-performance} 
    \vspace{-3mm}
\end{table}

\section{Methodology}\label{sec:method}

In order to design an effective and efficient method for mitigating the inherent positional bias of pre-trained LLMs, we introduce a strategy named \textbf{P}osition \textbf{A}ware \textbf{P}arameter \textbf{E}fficient \textbf{F}ine \textbf{T}uning (\textbf{PAPEFT}). It combines a position-aware parameter efficient adapter module with data augmentation techniques. Specifically, to remove the model's intrinsic location preference bias, which is typically introduced by the pre-training phase data, we employ a data augmentation technique (Section~\ref{section:method-train}) that involves random permutation of the ordering in candidate lists. This requires LLMs to distribute their attention uniformly across different positions within the input context. Furthermore, to efficiently debias the original parameters of LLMs, we introduce an new adapter module that explicitly incorporates the positional context of each candidate as learnable soft prompts (Section~\ref{section:method-1}). This integration aims to adjust the LLM's attention to various positions more equitably without modifying the original pre-trained parameters. 


\subsection{Ordering Permutation with Data Augmentation}\label{section:method-train}
As discussed in Section~\ref{section:empirical}, existing pre-trained LLMs exhibit a specific location preference over the input contextual length. This tendency results in an uneven distribution of attention across the entire context, and thus leads to fluctuating performance. A potential explanation for the distinct location preferences observed in various LLMs could stem from positional biases present in the original pre-training data, e.g. key information is often placed at the start of text. 

In order to provide an appropriate way to mitigate this issue in the data perspective, we adopt a strategic data augmentation process designed to evenly distribute LLM attention across various positions within the input context. Specifically, this approach creates multiple permutations for each set of potential document candidates within a given input context. These permutations serve as augmented fine-tuning datasets. Mathematically, given a list of candidates $[\mathbf{P}_s, \mathbf{X}_1, \mathbf{X}_2, \dots, \mathbf{X}_K]$, we create multiple permutations of this list, ensuring that each candidate $\mathbf{X}_i$ occupies every possible position across different versions. This generates a series of new training instances $[\mathbf{P}_s, \mathbf{X}_{\pi(1)}, \mathbf{X}_{\pi(2)}, \dots, \mathbf{X}_{\pi(K)}]$, where $\pi$ denotes a permutation function that rearranges the indices $1, 2, \dots, K$. 

\begin{figure}[t]
    \centering
    \includegraphics[width=.7\textwidth]{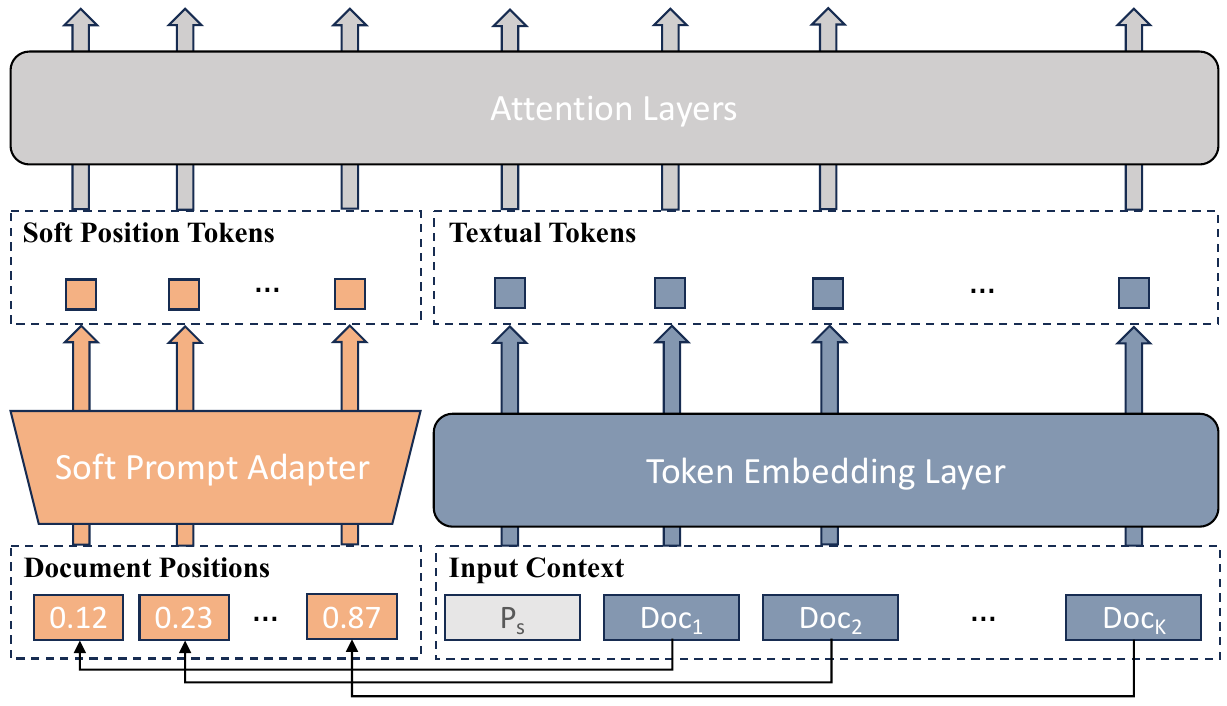} 
    \vspace{-5mm}
\caption{The overall framework of location encoding soft prompt adapter module. The relative locations of potential documents are initially computed and subsequently fed into a soft prompt adapter. The soft location tokens are concatenated with textual tokens to form a combined input for the attention layers in LLMs.}
\label{fig:framework-1} 
\vspace{-5mm}
\end{figure}

\subsection{Explicitly Incorporating Positions Location through Location Encoding Adapter}\label{section:method-1}
Given the generated augmented data, a key question is the design of the fine-tuning module for the pre-trained LLM. Directly optimizing the LLM parameters is a straightforward approach but proves inefficient due to the vast number of parameters involved. 
Although various parameter-efficient fine-tuning methods like LoRA~\citep{hu2021lora}, QLoRA~\citep{dettmers2023qlora}, and prompt tuning~\citep{lester2021power} are available, they do not adequately consider the location of documents within the input context, thus these approaches are not fully optimized in addressing positional bias.


In order to let LLMs be aware of the position of all potential documents for a debias-oriented optimization process, we further propose a novel adapter module to explicitly incorporate the relative locations of documents as additional input prompts, which is named as \underline{L}ocation \underline{E}ncoding (LE) adapter. Specifically, each document's relative location is computed and included in the input prompts. Then a trainable adapter module transforms the dimensions of these locational prompts into the token embedding space, aligning the semantic meaning of the transformed locational tokens with the textual tokens.


Mathematically, as illustrated in Figure~\ref{fig:framework-1}, the process begins by computing the relative locations of all potential documents within the context length, denoted as $\mathbf{S}\in \mathbb{R}^{K}$. A learnable adapter module $f_{\theta}$, where $\theta$ denotes the learnable parameters, is then applied to these locational prompts. This module aligns the semantic essence of these spatial tokens with the trained textual token space of the LLM.
The formal transformation process is represented as follows:
\begin{equation}
f_{\theta}(\mathbf{S}_i) = \mathbf{A}_i, \quad \forall i \in [1, K], \quad \mathbf{A}_i \in \mathbb{R}^{d},
\end{equation}
where $d$ represents the dimension of the token embedding space utilized by the LLM. 

Upon completion of this mapping process, the adapter model generates additional transformed tokens $\mathbf{A}$. These tokens are subsequently concatenated with the original textual tokens, forming a new, enriched input sequence $I = [\mathbf{A}, \mathbf{P}_s, \mathbf{X}_1, \mathbf{X}_2, \dots, \mathbf{X}_K]$ for the LLMs. This concatenated sequence encourages that each document is presented and guided by a contextual positional cue, thereby providing the LLM with a dual awareness of content and contextual positioning.

\vspace{-2mm}\section{Experiments}\label{sec:results}\vspace{-1mm}

\begin{table}[t]
\centering 
\begin{adjustbox}{width=1.0\columnwidth,center}
\begin{tabular}{@{}c|c|lcccccc|cc@{}}
\toprule
\multirow{2}{*}{Task} & \multirow{2}{*}{Base Model}          & \multicolumn{1}{c}{\multirow{2}{*}{Fine Tune Strategy}} & \multirow{2}{*}{1}     & \multirow{2}{*}{5}     & \multirow{2}{*}{9}     & \multirow{2}{*}{13}    & \multirow{2}{*}{17}    & \multirow{2}{*}{20}    & \multirow{2}{*}{Mean} & \multirow{2}{*}{Fluctuation (\%)} \\
                      &                                     &                                                         &       &       &       &       &       &       &                           &                              \\ 
\midrule
\multirow{8}{*}{REC} & \multirow{4}{*}{Longchat-13b-16k}    & Original                                                 & 0.329 & 0.249 & 0.211 & 0.205 & 0.171 & 0.341 & 0.251                     & 27.78                        \\
                      &                                      & PAPEFT-PT                                            & 0.832 & 0.714 & 0.708 & 0.723 & 0.715 & 0.736 & 0.733                     & 6.38                         \\
                      &                                      & PAPEFT-LE                                                    & 0.854 & 0.731 & 0.748 & 0.745 & 0.767 & 0.752 & 0.766                     & 5.82                         \\ 
                      &                                      & PAPEFT-LoRA       & 0.864 & 0.816 & 0.808 & 0.823 & 0.815 & 0.836 & 0.827   & 2.47        \\ \cmidrule(lr){2-11}
                      & \multirow{4}{*}{Vicuna-13b-v1.5-16k} & Original                                                 & 0.855 & 0.083 & 0.211 & 0.205 & 0.171 & 0.341 & 0.311                     & 89.76                        \\
                      &                                      & PAPEFT-PT                                           & 0.881 & 0.698 & 0.701 & 0.745 & 0.767 & 0.741 & 0.756                     & 8.88                         \\
                      &                                      & PAPEFT-LE                                                   & 0.883 & 0.746 & 0.738 & 0.798 & 0.807 & 0.765 & 0.790                     & 6.77                         \\ 
                      &                                      & PAPEFT-LoRA       & 0.855 & 0.836 & 0.818 & 0.833 & 0.825 & 0.855 & 0.837   & 1.83        \\
\midrule
\multirow{8}{*}{LP}  & \multirow{4}{*}{Longchat-13b-16k}    & Original                                                 & 0.016 & 0.112 & 0.147 & 0.168 & 0.051 & 0.022 & 0.086                     & 75.97                        \\
                      &                                      & PAPEFT-PT                                            & 0.698 & 0.708 & 0.742 & 0.760 & 0.718 & 0.742 & 0.728                     & 3.26                         \\
                      &                                      & PAPEFT-LE                                                    & 0.755 & 0.754 & 0.763 & 0.781 & 0.773 & 0.763 & 0.765                     & 1.37                         \\ 
                      &                                      & PAPEFT-LoRA        & 0.829 & 0.810 & 0.815 & 0.809 & 0.816 & 0.825 & 0.817   & 0.99 \\  \cmidrule(lr){2-11}
                      & \multirow{4}{*}{Vicuna-13b-v1.5-16k} & Original                                                 & 0.257 & 0.208 & 0.119 & 0.166 & 0.096 & 0.104 & 0.158                     & 40.58                        \\
                      &                                      & PAPEFT-PT                                            & 0.757 & 0.721 & 0.709 & 0.741 & 0.761 & 0.771 & 0.743                     & 3.27                         \\
                      &                                      & PAPEFT-LE                                                    & 0.744 & 0.774 & 0.773 & 0.769 & 0.760 & 0.783 & 0.767                     & 1.77                         \\
                      &                                      & PAPEFT-LoRA       & 0.824 & 0.824 & 0.823 & 0.841 & 0.843 & 0.853 & 0.835   & 1.52 \\
\bottomrule
\end{tabular}
\end{adjustbox}
\vspace{-4mm}
    \caption{Accuracy score and performance fluctuation with varying position of relevant document. We present the results of the original model, and the proposed PAPEFT framework equiped with three different parameter efficient fine-tuning techniques (PT for prompt tuning, LE for location encoding, and LoRA) on recommendation (REC) and link prediction (LP) tasks, with Longchat-13b-16k and Vicuna-13b-v1.5-16k as the base model.}
    \label{fig:all}
\vspace{-4mm}
\end{table}

\subsection{Experimental Settings}

Our introduced PAPEFT framework is composed of two main components: the data ordering permutation augmentation technique, and the parameter-efficient fine-tuning (PEFT) module. For an in-depth evaluation of the PEFT module, we have chosen three different choices. This choice is designed to cover a broad spectrum of tunable parameters and to evaluate the effectiveness of our specially designed location encoding soft prompt adapter. To differentiate between these configurations, we designate the variant equipped with the location encoding soft prompt adapter as \textbf{PAPEFT-LE}. The variant employing a straightforward prompt tuning adapter~\citep{lester2021power}, which shares the same architectural framework as the location encoding soft prompt adapter but removes the input of relative document locations, is termed \textbf{PAPEFT-PT}. Lastly, the variant incorporating a LoRA~\citep{hu2021lora} adapter is referred to as \textbf{PAPEFT-LoRA}. Table~\ref{tab:num_tunable_para} displays the tunable parameters of the three adapters.


\vspace{-2mm}
\paragraph{Augmented Datasets Details.} 
During data augmentation phase, we generated five permutations of each document set for REC, and three for LP tasks according to the size of datasets. We select Longchat-13b-16k~\citep{longchat2023} and Vicuna-13b-v1.5-16k~\citep{vicuna2023} as the base model for their proficiency in handling long context windows.  Statistics information about the datasets size can be found in Appendix~\ref{appendix:exp} Table~\ref{tab:data_statistics} and Table~\ref{tab:ft-data-stat}.

\vspace{-2mm}
\paragraph{Implementation Details.} 
For efficient fine-tuning, we enabled 4-bit loading of model. The training was conducted with a sequence length of 16,384 tokens, padding enabled to match this length. The soft prompt adapter is featured with a two-layer MLP encoder of 1024 hidden size. For optimization, we employed the paged AdamW 32-bit optimizer with a cosine learning rate scheduler, setting the initial learning rate at $2e^{-4}$. The model underwent four epochs of training with mixed precision of BFloat16 and Float32. Flash attention 2~\citep{dao2022flashattention} is used for further acceleration. For LoRA adapter, we use $r=16$ as the setting. We use standard next-token prediction as our training objective. All experiments were done using eight NVIDIA A100-40GB GPUs. Code can be found in \url{https://anonymous.4open.science/r/llm_long_context-E9CF}.

\subsection{Effectiveness Results}
Our key findings on accuracy results of REC and LP tasks, as shown in Figure~\ref{fig:all}, our proposed PAPEFT framework and original models are summarized as follows:

\textbf{Positional Bias in Original Models}: The performance of both Longchat-13b-16k and Vicuna-13b-v1.5-16k models demonstrated significantly noticeable fluctuations, with each model exhibiting distinct patterns of variability across different tasks. These fluctuations are indicative of prevalent positional bias within the original models.

\textbf{Reduction in Positional Fluctuations}: The PAPEFT framework achieves a substantial reduction in positional bias, with an average decrease in performance variance of 54.19\% for recommendation tasks and 58.72\% for link prediction tasks. This improvement signifies that the integrated approach of data augmentation and position-aware fine tuning effectively guides the LLM to treat all candidates within the input context more evenly, thus mitigating positional bias. 

\textbf{Enhancement in Model Performance}: The PAPEFT framework yielded substantial improvements in model performance with an average increase of 57.3\% for the recommendation task and 64.4\% for the link prediction task compared to the original model. These improvements demonstrate PAPEFT's ability to not only reduce performance bias but also to enhance the model's task-specific effectiveness. 

\textbf{Efficacy in Location Encoding Soft Prompt Module}: 
Furthermore, when comparing PAPEFT-LE to the prompt tuning method—which lacks location encoding but has an equivalent number of tunable parameters—PAPEFT-LE achieves an additional average reduction in performance fluctuations of 1.54\% and achieves an average performance improvement of 3.1\% over the prompt tuning method. This highlights the benefits of integrating explicit document locations via the soft prompt tuning module, underscoring its effectiveness.

\textbf{Parameter Efficiency of Location Encoding Soft Prompt Module}:
As highlighted in Table~\ref{tab:num_tunable_para}, PAPEFT-LE utilizes 23.87 times fewer parameters compared to the PAPEFT-LoRA method. Despite this, the results demonstrate that PAPEFT-LE almost achieves comparable performance improvement and variance deduction to the PAPEFT-LoRA method, which highlighting the parameter efficiency of location encoding soft prompt adapter.


\begin{table}[t]
\vspace{-2mm}
\centering
\begin{adjustbox}{width=.5\columnwidth,center}
\begin{tabular}{lrc}
\toprule
\textbf{Task} & \# Para & Ratio (\%) to original model \\
\midrule
PAPEFT-LoRA & 125,173,760 & 0.95 \\
\midrule
PAPEFT-LE  & 5,250,048 & 0.04 \\
\midrule
PAPEFT-PT  & 5,250,048 & 0.04 \\
\bottomrule
\end{tabular}
\end{adjustbox}\vspace{-3mm}
\caption{Number of tunable parameters comparison.}
\label{tab:num_tunable_para}
\vspace{-5mm}
\end{table}
\section{Conclusion}
In this work, we conducted a comprehensive investigation into the phenomenon of positional bias in large language models across diverse tasks that require retrieving relevant knowledge. Through empirical results, we demonstrated that current LLMs exhibit a noticeable positional preference over the candidate lists. We showed that merely adopting prompt-based solution is insufficient to address the positional bias issue. In order to address the positional bias issue of LLMs, we introduced a data augmentation technique to permute the ordering of candidates within the textual context, and a position aware fine tuning module, which explicitly integrates the locational context of each document into the LLMs' input through a trainable adapter module. Our extensive experiments in recommendation and link prediction tasks demonstrate that the proposed module can substantially mitigate positional bias with limited tunable parameters.

\bibliography{colm2024_conference}

\begin{thebibliography}{27}
\providecommand{\natexlab}[1]{#1}
\providecommand{\url}[1]{\texttt{#1}}
\expandafter\ifx\csname urlstyle\endcsname\relax
  \providecommand{\doi}[1]{doi: #1}\else
  \providecommand{\doi}{doi: \begingroup \urlstyle{rm}\Url}\fi

\bibitem[Borgeaud et~al.(2022)Borgeaud, Mensch, Hoffmann, Cai, Rutherford, Millican, Van Den~Driessche, Lespiau, Damoc, Clark, et~al.]{borgeaud2022improving}
Sebastian Borgeaud, Arthur Mensch, Jordan Hoffmann, Trevor Cai, Eliza Rutherford, Katie Millican, George~Bm Van Den~Driessche, Jean-Baptiste Lespiau, Bogdan Damoc, Aidan Clark, et~al.
\newblock Improving language models by retrieving from trillions of tokens.
\newblock In \emph{International conference on machine learning}, pp.\  2206--2240. PMLR, 2022.

\bibitem[Chiang et~al.(2023)Chiang, Li, Lin, Sheng, Wu, Zhang, Zheng, Zhuang, Zhuang, Gonzalez, Stoica, and Xing]{vicuna2023}
Wei-Lin Chiang, Zhuohan Li, Zi~Lin, Ying Sheng, Zhanghao Wu, Hao Zhang, Lianmin Zheng, Siyuan Zhuang, Yonghao Zhuang, Joseph~E. Gonzalez, Ion Stoica, and Eric~P. Xing.
\newblock Vicuna: An open-source chatbot impressing gpt-4 with 90\%* chatgpt quality, March 2023.
\newblock URL \url{https://lmsys.org/blog/2023-03-30-vicuna/}.

\bibitem[Chowdhery et~al.(2023)Chowdhery, Narang, Devlin, Bosma, Mishra, Roberts, Barham, Chung, Sutton, Gehrmann, et~al.]{chowdhery2023palm}
Aakanksha Chowdhery, Sharan Narang, Jacob Devlin, Maarten Bosma, Gaurav Mishra, Adam Roberts, Paul Barham, Hyung~Won Chung, Charles Sutton, Sebastian Gehrmann, et~al.
\newblock Palm: Scaling language modeling with pathways.
\newblock \emph{Journal of Machine Learning Research}, 24\penalty0 (240):\penalty0 1--113, 2023.

\bibitem[Dao et~al.(2022)Dao, Fu, Ermon, Rudra, and R{\'e}]{dao2022flashattention}
Tri Dao, Dan Fu, Stefano Ermon, Atri Rudra, and Christopher R{\'e}.
\newblock Flashattention: Fast and memory-efficient exact attention with io-awareness.
\newblock \emph{Advances in Neural Information Processing Systems}, 35:\penalty0 16344--16359, 2022.

\bibitem[Dettmers et~al.(2023)Dettmers, Pagnoni, Holtzman, and Zettlemoyer]{dettmers2023qlora}
Tim Dettmers, Artidoro Pagnoni, Ari Holtzman, and Luke Zettlemoyer.
\newblock Qlora: Efficient finetuning of quantized llms.
\newblock \emph{arXiv preprint arXiv:2305.14314}, 2023.

\bibitem[Dettmers et~al.(2024)Dettmers, Pagnoni, Holtzman, and Zettlemoyer]{dettmers2024qlora}
Tim Dettmers, Artidoro Pagnoni, Ari Holtzman, and Luke Zettlemoyer.
\newblock Qlora: Efficient finetuning of quantized llms.
\newblock \emph{Advances in Neural Information Processing Systems}, 36, 2024.

\bibitem[Guu et~al.(2020{\natexlab{a}})Guu, Lee, Tung, Pasupat, and Chang]{guu2020retrieval}
Kelvin Guu, Kenton Lee, Zora Tung, Panupong Pasupat, and Mingwei Chang.
\newblock Retrieval augmented language model pre-training.
\newblock In \emph{International conference on machine learning}, pp.\  3929--3938. PMLR, 2020{\natexlab{a}}.

\bibitem[Guu et~al.(2020{\natexlab{b}})]{guu2020realm}
Kelvin Guu et~al.
\newblock Realm: Retrieval-augmented language model pre-training.
\newblock In \emph{Proceedings of ICML}, 2020{\natexlab{b}}.

\bibitem[Houlsby et~al.(2019)Houlsby, Giurgiu, Jastrzebski, Morrone, De~Laroussilhe, Gesmundo, Attariyan, and Gelly]{houlsby2019parameter}
Neil Houlsby, Andrei Giurgiu, Stanislaw Jastrzebski, Bruna Morrone, Quentin De~Laroussilhe, Andrea Gesmundo, Mona Attariyan, and Sylvain Gelly.
\newblock Parameter-efficient transfer learning for nlp.
\newblock In \emph{International conference on machine learning}, pp.\  2790--2799. PMLR, 2019.

\bibitem[Hu et~al.(2021)Hu, Shen, Wallis, Allen-Zhu, Li, Wang, Wang, and Chen]{hu2021lora}
Edward~J Hu, Yelong Shen, Phillip Wallis, Zeyuan Allen-Zhu, Yuanzhi Li, Shean Wang, Lu~Wang, and Weizhu Chen.
\newblock Lora: Low-rank adaptation of large language models.
\newblock \emph{arXiv preprint arXiv:2106.09685}, 2021.

\bibitem[Jin et~al.(2023)Jin, Mao, Li, Jiang, Luo, Wen, Han, Lu, Wang, Li, et~al.]{jin2023amazon}
Wei Jin, Haitao Mao, Zheng Li, Haoming Jiang, Chen Luo, Hongzhi Wen, Haoyu Han, Hanqing Lu, Zhengyang Wang, Ruirui Li, et~al.
\newblock Amazon-m2: A multilingual multi-locale shopping session dataset for recommendation and text generation.
\newblock \emph{arXiv preprint arXiv:2307.09688}, 2023.

\bibitem[Kwiatkowski et~al.(2019)Kwiatkowski, Palomaki, Redfield, Collins, Parikh, Alberti, Epstein, Polosukhin, Devlin, Lee, et~al.]{kwiatkowski2019natural}
Tom Kwiatkowski, Jennimaria Palomaki, Olivia Redfield, Michael Collins, Ankur Parikh, Chris Alberti, Danielle Epstein, Illia Polosukhin, Jacob Devlin, Kenton Lee, et~al.
\newblock Natural questions: a benchmark for question answering research.
\newblock \emph{Transactions of the Association for Computational Linguistics}, 7:\penalty0 453--466, 2019.

\bibitem[Lester et~al.(2021)Lester, Al-Rfou, and Constant]{lester2021power}
Brian Lester, Rami Al-Rfou, and Noah Constant.
\newblock The power of scale for parameter-efficient prompt tuning.
\newblock \emph{arXiv preprint arXiv:2104.08691}, 2021.

\bibitem[Lewis et~al.(2020)]{lewis2020retrieval}
Patrick Lewis et~al.
\newblock Retrieval-augmented generation for knowledge-intensive nlp tasks.
\newblock In \emph{Proceedings of NeurIPS}, 2020.

\bibitem[Li et~al.(2023)Li, Shao, Xie, Sheng, Zheng, Gonzalez, Stoica, Ma, and Zhang]{longchat2023}
Dacheng Li, Rulin Shao, Anze Xie, Ying Sheng, Lianmin Zheng, Joseph~E. Gonzalez, Ion Stoica, Xuezhe Ma, and Hao Zhang.
\newblock How long can open-source llms truly promise on context length?, June 2023.
\newblock URL \url{https://lmsys.org/blog/2023-06-29-longchat}.

\bibitem[Li \& Liang(2021)Li and Liang]{li2021prefix}
Xiang~Lisa Li and Percy Liang.
\newblock Prefix-tuning: Optimizing continuous prompts for generation.
\newblock \emph{arXiv preprint arXiv:2101.00190}, 2021.

\bibitem[Liu et~al.(2023{\natexlab{a}})Liu, Lin, Hewitt, Paranjape, Bevilacqua, Petroni, and Liang]{liu2023lost}
Nelson~F Liu, Kevin Lin, John Hewitt, Ashwin Paranjape, Michele Bevilacqua, Fabio Petroni, and Percy Liang.
\newblock Lost in the middle: How language models use long contexts.
\newblock \emph{arXiv preprint arXiv:2307.03172}, 2023{\natexlab{a}}.

\bibitem[Liu et~al.(2023{\natexlab{b}})Liu, Yuan, Fu, Jiang, Hayashi, and Neubig]{liu2023pre}
Pengfei Liu, Weizhe Yuan, Jinlan Fu, Zhengbao Jiang, Hiroaki Hayashi, and Graham Neubig.
\newblock Pre-train, prompt, and predict: A systematic survey of prompting methods in natural language processing.
\newblock \emph{ACM Computing Surveys}, 55\penalty0 (9):\penalty0 1--35, 2023{\natexlab{b}}.

\bibitem[Liu et~al.(2023{\natexlab{c}})Liu, Wang, Dao, Zhou, Yuan, Song, Shrivastava, Zhang, Tian, Re, et~al.]{liu2023deja}
Zichang Liu, Jue Wang, Tri Dao, Tianyi Zhou, Binhang Yuan, Zhao Song, Anshumali Shrivastava, Ce~Zhang, Yuandong Tian, Christopher Re, et~al.
\newblock Deja vu: Contextual sparsity for efficient llms at inference time.
\newblock In \emph{International Conference on Machine Learning}, pp.\  22137--22176. PMLR, 2023{\natexlab{c}}.

\bibitem[Min et~al.(2021)Min, Lewis, Zettlemoyer, and Hajishirzi]{min2021metaicl}
Sewon Min, Mike Lewis, Luke Zettlemoyer, and Hannaneh Hajishirzi.
\newblock Metaicl: Learning to learn in context.
\newblock \emph{arXiv preprint arXiv:2110.15943}, 2021.

\bibitem[Naumov et~al.(2019)Naumov, Mudigere, Shi, Huang, Sundaraman, Park, Wang, Gupta, Wu, Azzolini, et~al.]{naumov2019deep}
Maxim Naumov, Dheevatsa Mudigere, Hao-Jun~Michael Shi, Jianyu Huang, Narayanan Sundaraman, Jongsoo Park, Xiaodong Wang, Udit Gupta, Carole-Jean Wu, Alisson~G Azzolini, et~al.
\newblock Deep learning recommendation model for personalization and recommendation systems.
\newblock \emph{arXiv preprint arXiv:1906.00091}, 2019.

\bibitem[Ravaut et~al.(2023)Ravaut, Joty, Sun, and Chen]{ravaut2023position}
Mathieu Ravaut, Shafiq Joty, Aixin Sun, and Nancy~F Chen.
\newblock On position bias in summarization with large language models.
\newblock \emph{arXiv preprint arXiv:2310.10570}, 2023.

\bibitem[Roberts et~al.(2020)Roberts, Raffel, and Shazeer]{roberts2020much}
Adam Roberts, Colin Raffel, and Noam Shazeer.
\newblock How much knowledge can you pack into the parameters of a language model?
\newblock \emph{arXiv preprint arXiv:2002.08910}, 2020.

\bibitem[Vaswani et~al.(2017)]{vaswani2017attention}
Ashish Vaswani et~al.
\newblock Attention is all you need.
\newblock \emph{Advances in neural information processing systems}, 30, 2017.

\bibitem[Wang et~al.(2020)Wang, Shen, Huang, Wu, Dong, and Kanakia]{wang2020microsoft}
Kuansan Wang, Zhihong Shen, Chiyuan Huang, Chieh-Han Wu, Yuxiao Dong, and Anshul Kanakia.
\newblock Microsoft academic graph: When experts are not enough.
\newblock \emph{Quantitative Science Studies}, 1\penalty0 (1):\penalty0 396--413, 2020.

\bibitem[Yasunaga et~al.(2021)Yasunaga, Ren, Bosselut, Liang, and Leskovec]{yasunaga2021qa}
Michihiro Yasunaga, Hongyu Ren, Antoine Bosselut, Percy Liang, and Jure Leskovec.
\newblock Qa-gnn: Reasoning with language models and knowledge graphs for question answering.
\newblock \emph{arXiv preprint arXiv:2104.06378}, 2021.

\bibitem[Zheng et~al.(2023)Zheng, Zhou, Meng, Zhou, and Huang]{zheng2023large}
Chujie Zheng, Hao Zhou, Fandong Meng, Jie Zhou, and Minlie Huang.
\newblock On large language models' selection bias in multi-choice questions.
\newblock \emph{arXiv preprint arXiv:2309.03882}, 2023.

\end{thebibliography}
\bibliographystyle{colm2024_conference}

\appendix
\newpage
\section{Appendix}\label{appendix:exp}

In Table~\ref{tab:data_statistics}, we show the basic statistic information about the datasets used in experiments. In Table~\ref{tab:ft-data-stat}, we show the statistic information about the training and test datasets used in fine tune phase. 

\begin{table}[h]
\centering
\begin{tabular}{lrc}
\toprule
\textbf{Task} & \textbf{$K$} & \textbf{Average \# of Words} \\
\midrule
REC & 20 & 4.2k \\
\midrule
LP             & 20 & 6.2k \\
\bottomrule
\end{tabular}
\caption{Data statistics for test inference. Here $K$ denotes the number of potential items to select.}
\label{tab:data_statistics}
\end{table}

\begin{table}[h]
\centering
\begin{tabular}{lrr}
\toprule
\textbf{Task} & \# Train & \# Test \\
\midrule
REC & 2,000 & 1,000 \\
\midrule
LP             & 10,000 & 3,000 \\
\bottomrule
\end{tabular}
\caption{  Fine-tune data train test splits statistics. }
\label{tab:ft-data-stat}
\end{table}

\begin{table*}[t] \centering
\caption{Examples of zero-shot prompts used in different domains.}
\label{zeroshot-prompt}
\begin{tabular}{p{0.6cm}p{13.0cm}}\toprule
Task \newline setting & Prompt to LLM\\ \midrule
\textsc{Rec} &  Task: Using a user's historical purchase data from Amazon.com, identify one product from a distinct list of potential products that you predict the user will most likely purchase next. \newline\newline Belows are 2 historical purchased products: \newline Bought Product [1](Title: New brothread Wash Away)\newline Bought Product [2](Title: Crafter's Companion Spray \& Shine, Varnish) \newline\newline Belows are 20 potential products to consider:\newline \textbf{Potential Product [1](Title: Clarins Eau Dynamisante Shower Gel 150ml)}\newline Potential Product [2](Title: My Living World LW105 Window Bird Feeder) \newline \dots \newline Potential Product [20](Title: BGS Do it yourself| Cutting Box with Fine Saw)\newline Question: Now you need to predict ONLY one product from the potential products that the user will most likely purchase next. What is your prediction: \\ \hline
\textsc{LP} & Task: Based on the title and abstract of a research paper, determine one paper from a list of potential papers that the original paper is most likely to cite.\newline\newline
Below is the provided research paper along with its title and abstract: style aggregated network for facial landmark detection (Abstract:  …)
\newline
The following are 20 potential papers for consideration:

Potential Paper [1](decafa deep convolutional cascade for face alignment in the wild) (Abstract:  …)

Potential Paper [2](do altmetrics work for assessing research quality ) (Abstract:  …)

…

\textbf{Potential Paper [20](ai based pilgrim detection using convolutional neural networks ) (Abstract:  …)}

Question: Predict ONE paper from the given potential papers that the original document would most probably cite.

Please provide the predicted paper and a brief description of why you think it is the most likely choice:
\\
\bottomrule
\end{tabular}
\end{table*}

\begin{table*}[t] \centering
\caption{Prompt engineering examples, few-shot learning and hierarachical settings.}
\label{prompt-engineering}
\begin{tabular}{p{1.2cm}p{12.0cm}}\toprule
Prompt \newline strategy & Prompt to LLM\\ \midrule
\textsc{FEW-shot} &  Task: \textcolor{black}{[Task Description]} \newline\newline Belows are 3 examples: \newline Example [1] \textcolor{blue}{[Exampel 1]} \newline Example [2] \textcolor{blue}{[Exampel 2]} \newline Example [3] \textcolor{blue}{[Exampel 3]} \newline\newline Belows are 2 historical purchased products: \newline Bought Product [1]\textcolor{black}{[Historical Bought Product 1]} \newline Bought Product [2]\textcolor{black}{[Historical Bought Product 2]} \newline\newline Belows are 20 potential products to consider:\newline \textbf{Potential Product [1] \textcolor{black}{[Potential Product 1]}}\newline Potential Product [2] \textcolor{black}{[Potential Product 2]} \newline \dots \newline Potential Product [20] \textcolor{black}{[Potential Product 20]}\newline Question: Now you need to predict ONLY one product from the potential products that the user will most likely purchase next. What is your prediction: \\ \hline
\textsc{Hierar-chical} &  Task: \textcolor{black}{[Task Description]} \newline\newline \textcolor{blue}{Given the inherent challenge of selecting the prime candidate directly from a broad list, approach this assignment hierarchically: Start by segmenting the products into 5 equal groups. For each segmented group, determine the product with the highest purchase likelihood. For instance, select the most likely one from group ([1]-[4]), followed by the top pick from group ([5]-[8]), and so on. After narrowing down to the top products from each group, decide which among them stands the best chance of being the user's next purchase.} \newline\newline Belows are 2 historical purchased products: \newline Bought Product [1]\textcolor{black}{[Historical Bought Product 1]} \newline Bought Product [2]\textcolor{black}{[Historical Bought Product 2]} \newline\newline Belows are 20 potential products to consider:\newline \textbf{Potential Product [1] \textcolor{black}{[Potential Product 1]}}\newline Potential Product [2] \textcolor{black}{[Potential Product 2]} \newline \dots \newline Potential Product [20] \textcolor{black}{[Potential Product 20]}\newline Question: Now you need to predict ONLY one product from the potential products that the user will most likely purchase next. What is your prediction: 
\\
\bottomrule
\end{tabular}
\end{table*}
\end{document}